\pdfoutput=1

\documentclass[11pt]{article}

\usepackage[final]{acl}

\usepackage{times}
\usepackage{latexsym}
\usepackage{amsmath}
\usepackage{booktabs}

\usepackage[T1]{fontenc}

\usepackage[utf8]{inputenc}

\usepackage{microtype}

\usepackage{inconsolata}

\usepackage{graphicx}

\usepackage[utf8]{inputenc} %
\usepackage[T1]{fontenc}    %
\usepackage{hyperref}       %
\usepackage{url}            %
\usepackage{booktabs}       %
\usepackage{amsfonts}       %
\usepackage{nicefrac}       %
\usepackage{microtype}      %
\usepackage{xcolor}         %

\usepackage{amsmath}
\usepackage{amssymb}
\usepackage{mathtools}
\usepackage{amsthm}
\usepackage{tcolorbox}
\usepackage[capitalize,noabbrev]{cleveref}
\usepackage{pgfplots}
\pgfplotsset{compat=1.18}
\usepackage{multirow}
\theoremstyle{plain}

\theoremstyle{definition}

\theoremstyle{remark}

\usepackage[textsize=tiny]{todonotes}
\usepackage{mathrsfs}
\usepackage{times}
\usepackage{latexsym}
\usepackage[T1]{fontenc}
\usepackage[utf8]{inputenc}
\usepackage{microtype}
\usepackage{inconsolata}

\usepackage{CJKutf8}
\usepackage{times}
\usepackage{latexsym}
\usepackage{graphicx}
\usepackage{makecell}
\usepackage{hyperref}
\usepackage{wrapfig}
\usepackage{subfig}
\usepackage{multicol}
\usepackage{multirow}
\usepackage{subcaption}

\title{Stable Language Model Pre-training by Reducing Embedding Variability}

\author{
  Woojin Chung \hspace{0.3cm} Jiwoo Hong \hspace{0.3cm} Na Min An \hspace{0.3cm} James Thorne \thanks{Corresponding author} \hspace{0.3cm} Se-Young Yun \footnotemark[1] \\[0.5cm]
  \large KAIST AI\\
  \texttt{\{gartland, jiwoo\_hong, naminan, thorne, yunseyoung\}@kaist.ac.kr}
}

\begin{document}
\maketitle

\begin{abstract}

Stable pre-training is essential for achieving better-performing language models. However, tracking pre-training stability by calculating gradient variance at every step is impractical due to the significant computational costs. We explore Token Embedding Variability (TEV) as a simple and efficient proxy for assessing pre-training stability in language models with pre-layer normalization, given that shallower layers are more prone to gradient explosion (section \ref{subsec:stability and TEV}). Moreover, we propose Multi-head Low-Rank Attention (MLRA) as an architecture to alleviate such instability by limiting the exponential growth of output embedding variance, thereby preventing the gradient explosion (section \ref{sec:theory}). Empirical results on GPT-2 with MLRA demonstrate increased stability and lower perplexity, particularly in deeper models.
\end{abstract}

\section{Introduction}
\label{sec:intro}

Improving large language models (LLMs) typically involves increasing model size, especially through greater depth \citep{brown2020language, kaplan2020scaling, rae2022scaling, xue2023study}. However, this approach often causes instability during pre-training, indicated by sudden spikes in loss \citep{chowdhery2022palm, zhai2023stabilizing}, while stable pre-training typically leads to stronger performance under controlled training configurations \cite{touvron2023llama, takase2024spike}. Such instability can lead to catastrophic divergence or degradation, underscoring the importance of assessing pre-training stability \citep{chowdhery2022palm, zhai2023stabilizing, takase2024spike}.

\begin{figure}[t!]
    \centering
    \includegraphics[width=\columnwidth]{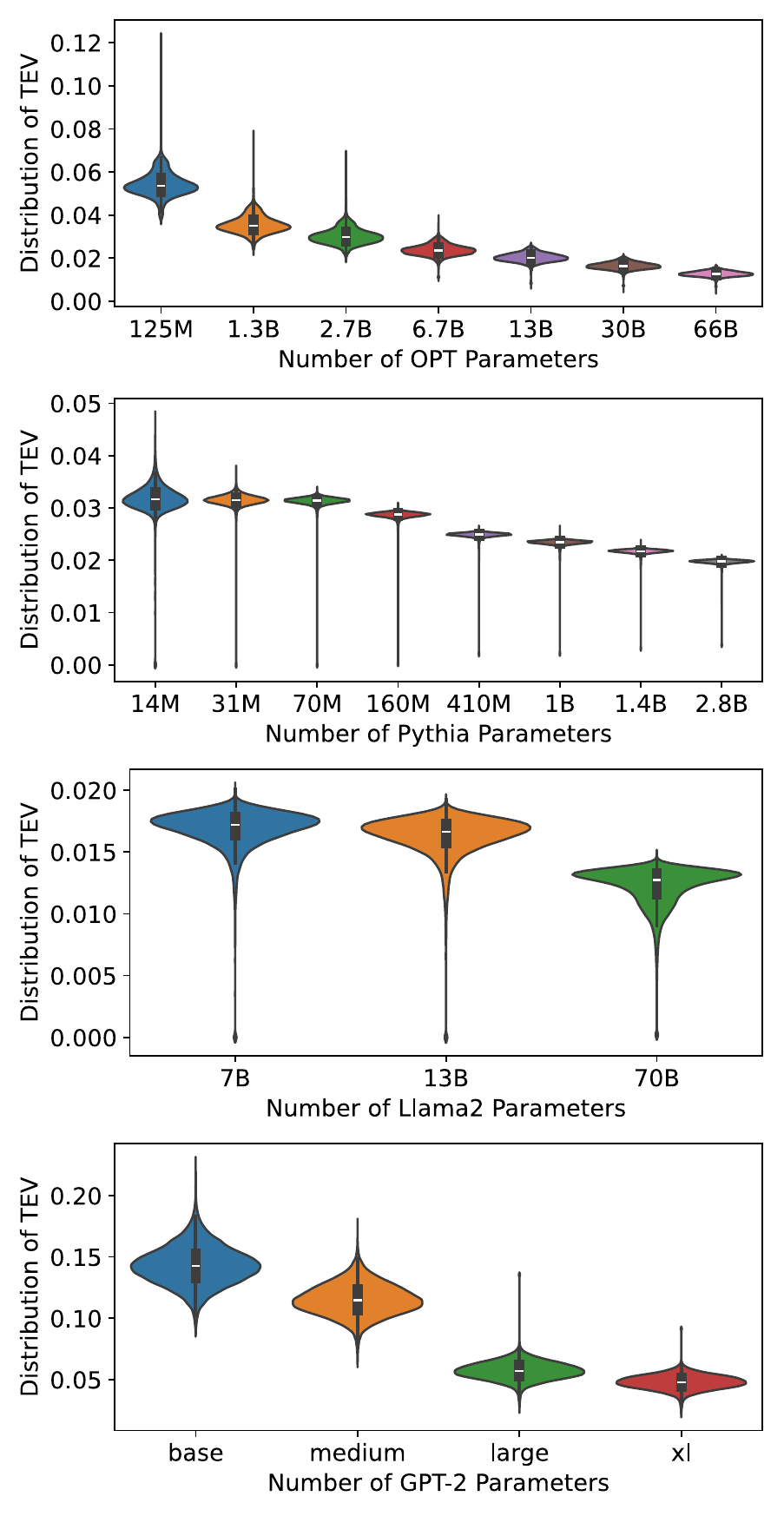}
    \caption{TEV distribution for OPT, Pythia, Llama-2, and GPT-2 reveals that as model size grows, both $\mu_{\text{TEV}}$ and $\sigma_{\text{TEV}}$ decrease. This trend correlates with better model performance, as reduced noisy gradients lead to higher pre-training stability and improved performance. For a fair comparison, Pythia 6.9B and 12B were excluded due to their different vocabulary sizes.}
    \label{fig:tev_models}
    \vspace{-0.2in}
\end{figure}

The conventional methods for monitoring the pre-training stability are computationally expensive \cite{kaplan2020scaling}, such as observing the gradient variance which needs additional $O(nd)$ for gradient matrix $g_t \in \mathbb{R}^{n \times d}$ \citep{zhao2024galore} or analyzing the singular values of the second-order derivative of the loss with respect to model parameters \citep{yao2020pyhessian,gilmer2021loss,cohen2024adaptive}. Further details will be addressed in Appendix \ref{apdx:related}.

We address both issues by dissecting the token embedding layer: we theoretically and empirically substantiate that the standard deviation of token embedding in the embedding layer, denoted token embedding variability (TEV), can be a simple and efficient proxy for estimating pre-training stability in models with pre-layer normalization \cite{radford2019language, zhang2022opt, touvron2023llama2} as it best reflects the level of gradient noise (\emph{i.e.,} gradient variance). We demonstrate a correlation between TEV and language model performance by evaluating OPT \cite{zhang2022opt}, Pythia \cite{biderman2023pythia}, Llama-2 \cite{touvron2023llama2}, and GPT-2 \cite{radford2019language} (Figure~\ref{fig:tev_models}). Furthermore, we introduce factorized multi-head attention projection matrices (\emph{i.e.,}, Multi-head Low-Rank attention; MLRA) as a fundamental method to mitigate pre-training instability. we empirically show that pre-training GPT-2 \citep{radford2018improving} with MLRA effectively \textit{lowers} TEV and achieves higher downstream performance with better pre-training stability, aligning with the theoretical analysis of TEV.

\section{Pre-training Stability Proxy} \label{sec:tev}
\subsection{Preliminaries}

The token embedding layer $\mathbf{E} \in \mathbb{R}^{|V| \times d_{\text{model}}}$ of the transformer \citep{vaswani2023attention} maps an input sequence $\mathbf{x} = \left[ x_1, x_2, \cdots, x_n \right]$ with $n$ tokens into the vector-wise representations $X_0 \in \mathbb{R}^{n \times d_{\text{model}}}$,
\begin{equation*}
    \mathbf{E} = \begin{bmatrix}
                    \mathbf{e}_1 & \mathbf{e}_2 & \cdots & \mathbf{e}_{|V|}
                 \end{bmatrix}^\textnormal{T},
\end{equation*}
where $|V|$ and $ d_{\text{model}}$ refer to the size of vocabulary and the hidden dimension, and $e_i \in \mathbb{R}^{d_{\text{model}}}$ denotes the embedding weight vector corresponding to each token. Thus, $\textbf{e}_i$ can be written as:
\begin{equation*}
    \textbf{e}_i = \begin{pmatrix}
            e_{i,1} & e_{i,2} & \cdots & e_{i,d_{\text{model}}}
          \end{pmatrix}.
\end{equation*}
The initial embedding vectors, $X_0 \in \mathbb{R}^{n \times d_{\text{model}}}$, pass through $2N$ different sub-layers $\mathcal{F}$, 
\begin{equation*}
    X_t = \mathcal{F}_t(X_{t-1}) + X_{t-1} 
\end{equation*}
where $t \in \{ 1, 2, \ldots, N \}$ denotes the layer index, $X_t$ denotes the hidden representation returned from $t$-th layer. Finally, the logit $L \in \mathbb{R}^{n \times |V|}$ for predicting the next token is calculated by mapping $X_N$ into $|V|$-dimensional space with the language model head. Language models such as BERT \citep{devlin2019bert} GPT-2 \citep{radford2019language} Mistral \citep{jiang2023mistral} and Llama-2 \citep{touvron2023llama} typically tie language modeling head with the embedding matrix \( \mathbf{E} \) to reduce the number of trainable parameters and induce input and output embedding behaves similarly to similar words \cite{mnih2012fastsimplealgorithmtraining,press2017usingoutputembeddingimprove, inan2017tyingwordvectorsword}.
\begin{equation*}
    L = X_N \cdot \mathbf{E}^T.
\end{equation*}

\subsection{Stability and Token Embedding Layer} \label{subsec:stability and TEV}
We show that the token embedding layer $\mathbf{E}$ plays a crucial role in understanding the pre-training stability in two perspectives: 1) gradient explosion and 2) skewness in token frequency. %

\paragraph{Gradient explosion} Recently proposed LLMs typically apply pre-layer norm \citep[pre-LN]{prelayernorm} to mitigate pre-training instability in the early stage of pre-training due to high gradient variance (\emph{i.e.,} noisy gradient) \cite{liu2021varianceadaptivelearningrate} \footnote{Gradient mean close to $0$ in the early stage of pre-training as weights are initialized from normal distributions with mean $0$ \cite{balduzzi2018shatteredgradientsproblemresnets}. Exponential moving average amplifies variance of gradient estimation \cite{liu2021varianceadaptivelearningrate}}. Contrary to post-layer norm \citep[post-LN]{ba2016layernormalization,vaswani2023attention}, the gradient norms are usually larger in shallower layers compared to deeper layers \cite{xie2023residual}, leading the gradient of token embedding layer $\nabla X_0$ to have the greatest magnitude:
\begin{equation*}
    \nabla X_0 =  \nabla X_N \cdot \prod_{t=1}^{N-1} \left(\frac{\partial \mathcal{F}_{t-1}\left( X_{t-1} \right)}{\partial X_{t-1}} + \mathbf{I}\right),
\end{equation*}
as the gradient exponentially grows over the layers due to the residual connection \citep{he2016deep}. Such property, which causes spikes in pre-training loss, is amplified in the token embedding layer (\emph{i.e.,} gradient explosion). Thus, the token embedding layer $\mathbf{E}$ effectively reflects the training instability. We empirically confirm this in Section \ref{subsec:results}. For simplicity, we assume a negligible or zero correlation between the gradient and weight matrix ($\mathrm{Cov}(X_0, \nabla X_0) \approx 0$), and our experiment supports that this assumption is valid in real scenarios.
\begin{equation*}
\begin{aligned}
\mathrm{Var}(X_0 - \nabla X_0) &= \mathrm{Var}(X_0) + \mathrm{Var}(\nabla X_0) \\
&\quad - 2\mathrm{Cov}(X_0, \nabla X_0)
\end{aligned}
\end{equation*}

\paragraph{Skewness in token frequency} Since the true distribution of natural language is inherently non-uniform \citep{zipf1935}, mini-batch gradient descent leads to imbalanced updates of token embeddings. The gradient of the mini-batch is normalized by its total number of tokens \cite{laurent2024feature, dettmers20228bit}. The tokens in each mini-batch $B$ can be written as:
\begin{equation*}
    B = \left\{ \mathbf{x}_i \right\}_{i=1}^M = \left\{ \left[ x_{i,j} \right]_{j=1}^C \mid i = 1, 2, \ldots, M \right\},
\end{equation*}
where $M$ is the batch size and $C$ is the sequence length of each token. This could be understood as sampling a total of $M \times C$ independent random samples from the population $V$ \emph{with replacemenț}. Therefore, the skewed token distribution and mini-batch updates lead to the selective update of certain token's embedding weights of $\mathbf{E}$.

\subsection{Token Embedding Variability (TEV)}

When pre-training is stable, the norm of each token's embedding weight vector $||e_i||$ should be close to uniform. $||e_i||$ can be written as:
\begin{equation*}
    ||e_i|| = \sqrt{d_{\text{model}} \cdot \left( \mu_i^2 + \sigma_i^2 \right)}
\end{equation*}
where $\mu_i^2$ and $\sigma_i^2$ are element-wise mean and variance of $e_i$. $d_{\text{model}}$ is a fixed value with a positive integer, and $\mu_i$ stays close to zero throughout pre-training\footnote{Token embedding layer $\mathbf{E}$ is initialized using a normal distribution with a mean of zero.}. 
We confirmed that $\mu_i$ is close to zero in multiple pre-trained LLMs in Appendix \ref{apdx:mean}. Hence, $\sigma_i^2$ is the dominant term determining $||e_i||$. 

However, the standard deviation is typically significantly less than one, and the token embedding norm falls short as a reliable proxy of pre-training stability. Given that the model dimension (\(d_{\text{model}}\)) is a positive integer and generally much larger than the standard deviation (\(\sigma\)), the token embedding norm largely overlooks the standard deviation. This oversight is critical, as the standard deviation is key to capturing gradient variance during pre-training, which the norm fails to account for accurately.

\begin{figure*}[t!]
    \centering
    \includegraphics[width=\textwidth]{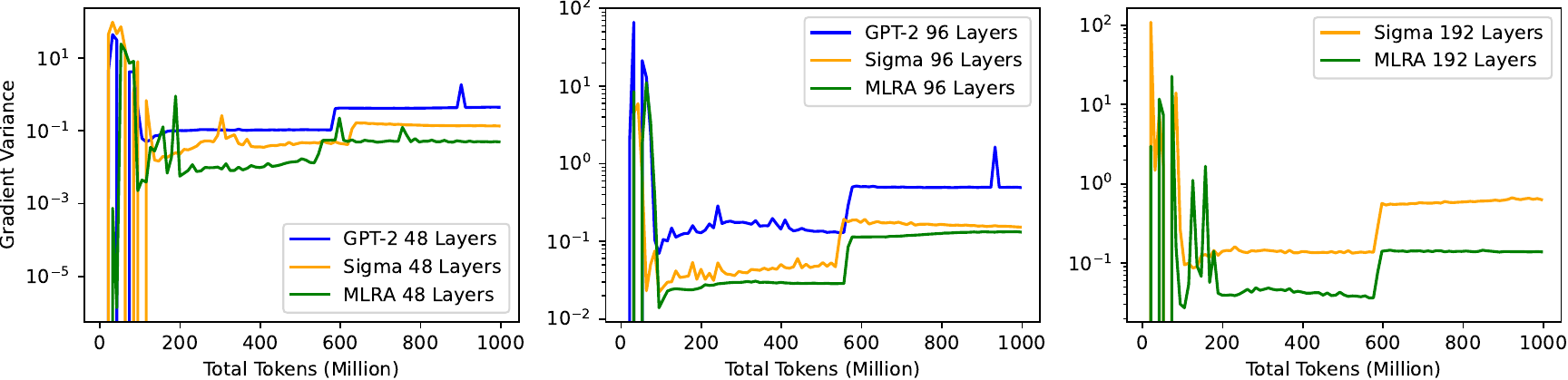}
    \caption{Gradient variance ($\downarrow$) comparison across tested models with different layers. MLRA shows the lowest gradient variance than GPT-2 and $\sigma$Reparam. GPT-2 with 192 layers was excluded as the training failed 5 times (\textit{i.e.}, The gradient variance is infinite at the earlier steps and becomes infinitesimal in the later steps).}
    \label{fig:grad_var}
\end{figure*}

Therefore, we propose the distribution of token-level standard deviation ($\sigma$) as the pre-training stability proxy: \emph{i.e.,} token embedding variability (TEV) distribution. TEV of $i$-th token ($x_i$) is defined as:
\begin{equation*}
    \text{TEV}_i = \sqrt{\frac{1}{d} \sum_{j=1}^{d} \left( e_{ij} - \bar{e}_i \right)^2},
\end{equation*}
where \( \bar{e}_i \) is the element mean of the $i$th token's weight vector. Eventually, the mean $\mu_{\text{TEV}}$ and standard deviation $\sigma_{\text{TEV}}$ of TEV over the entire vocabulary is:
\begin{align*}
    \mu_{\text{TEV}} &= \frac{1}{|V|} \sum_{i=1}^{|V|} \text{TEV}_i\\
    \sigma_{\text{TEV}} &= \sqrt{\frac{1}{|V|} \sum_{i=1}^{|V|} (\text{TEV}_i - \mu_{\text{TEV}})^2}
\end{align*}

Our experiments in Section \ref{subsec:results} verify that stable pre-training with less suffer from noisy gradient results in a TEV distribution with a lower $\mu_{\text{TEV}}$ and  $\sigma_{\text{TEV}}$.

\section{Mitigating TEV with Factorization} \label{sec:mlra}

We propose low-rank factorized attention projection matrices (\emph{i.e.,} Multi-head Low-Rank attention; MLRA) as a simple way of lowering TEV mean and variances, improving pre-training stability and performance. 

\subsection{Multi-head Low Rank Attention (MLRA)}

Query, key, and value projection matrices $W_q$, $W_k$ and $W_v$ can be factorized as:
\(WX_t = W^{U}W^{D}X_t\),
where $W \in \mathbb{R}^{d_{\text{model}} \times d_{\text{model}}}$, $W^{U} \in \mathbb{R}^{d_{\text{model}} \times r}$ and $W^{D} \in \mathbb{R}^{r \times d_{\text{model}}}$ $(r < d_{\text{model}})$, and $r$ refers to the rank. MLRA introduces minimal overhead since MLRA is only applied to the weights within the multi-head attention mechanism.
As $W$ can be reconstructed from the two low-rank matrices, there is no additional cost at inference.

\subsection{Theoretical Analysis} \label{sec:theory}

The factorization property of MLRA mitigates the exponential growth of variance in the output representations across layers. The variance with MLRA with the hidden representation in $t$th layer can be simply written as:
\begin{equation*}
\begin{split}
    \sigma^2(W^{U} W^{D}X_t) = r \cdot d_{\text{model}} \cdot \sigma^2(W^{U}) \cdot \sigma^2(W^{D}),
\end{split}
\label{eq:var}
\end{equation*}

assuming $X_t$, $W^{U}$ and $W^{D}$ are independent each other, and $X_t$ has zero mean. This is due to the independent initialization of weights from the identical distribution and the application of layer normalization to the input, guaranteeing a zero mean. Similarly, we can assume \(\sigma^2(X_t) = 1\).

For \(\sigma^2({W}^{U})\) and \(\sigma^2({W}^{D})\), we use the actual values from \texttt{torch.nn.Linear}, where the weights \( w \) are initialized using Kaiming uniform initialization \citep{he2015delving}, \( w \sim \mathcal{U}\left(-\sqrt{\frac{1}{n}}, \sqrt{\frac{1}{n}}\right) \) with \(\sigma^2(W) = \frac{1}{3d_{\text{model}}}\), \( W \in \mathbb{R}^{d_{\text{model}} \times d_{\text{model}}} \) (\emph{See} Appendices \ref{apdx:kaiming} for details). First, the initial variances of a square matrix attention weight and MLRA are \footnote{For $X_t \sim \mathcal{N}(0,1)$, and $W^U, W^D \sim \mathcal{N}(0, \sigma^2)$, $\sigma^2(WX_t) = d_{\text{model}} \cdot \sigma^2$ and $\sigma^2(W^U W^D X_t) = r \cdot d_{\text{model}} \cdot \sigma^4$. Weights are initialized with $\sigma = 0.02$ in huggingface library}:
\begin{align*}
    \sigma^2(WX_t) &= d_{\text{model}} \cdot \frac{1}{3d_{\text{model}}} = \frac{1}{3}\\
    \sigma^2(W^{U}W^{D}X_t) &= d_{\text{r}} \cdot d_{\text{model}} \cdot \frac{1}{3d_{\text{model}}} \cdot \frac{1}{3d_{\text{r}}} = \frac{1}{9}.
\end{align*}
Therefore, passing through two linear layers further diminishes the variance of the token embeddings. We further the extended calculation of variance growth of each attention head and self-attention in Appendix \ref{apdx:extension}. 

MLRA addresses gradient explosion in a similar manner to scaled initialization by reducing the magnitude of weights in both the feed-forward network and self-attention module during the weight initialization \cite{shoeybi2020megatronlm, scao2022language, biderman2023pythia, takase2024spike}. Unlike scaled initialization, MLRA uses standard initialization, leading to larger gradient updates.

On the other hand, simply applying low-rank reparameterization to all weights from the beginning of pre-training degrades performance due to the high intrinsic rank of weight matrices \cite{aghajanyan2020intrinsic, lialin2023stack, zhao2023inrank, zhao2024galore}. 
To address this, we concentrate on the multi-head architecture, which divides output representation across hidden dimensions, to mitigate low-rank bottlenecks. One example is as follows: Let matrix $\mathbf{A}$ be a $3 \times 6$ matrix with 
\[
\mathbf{A} = [\mathbf{e}_1, \mathbf{e}_2, \mathbf{e}_3, \mathbf{e}_1 + \mathbf{e}_2, \mathbf{e}_1 + \mathbf{e}_3, \mathbf{e}_2 + \mathbf{e}_3]
\]

where $\mathbf{e}_1, \mathbf{e}_2$ and $\mathbf{e}_3$ are the standard basis vectors in $\mathbb{R}^{3}$. The submatrices $\mathbf{A}_1 = [\mathbf{e}_1, \mathbf{e}_2, \mathbf{e}_3]$ and $\mathbf{A}_2 = [\mathbf{e}_1 + \mathbf{e}_2, \mathbf{e}_1 + \mathbf{e}_3, \mathbf{e}_2 + \mathbf{e}_3]$ both possess a rank of 3, illustrating that a rank-3 matrix $\mathbf{A}$ can still have full-rank submatrices, even when the matrix is divided along hidden dimensions. Thus we hypothesize that matrix factorization within a multi-head architecture could reduce gradient variance and avoid low-rank bottlenecks during pre-training.

\begin{table*}[t!]
\vskip 0.15in
\begin{center}
\begin{small}
\begin{sc}
\begin{tabular}{l|c|c|c|ccccccc}
\toprule
\textbf{Models} & \textbf{Layers} & \textbf{$\mu_{\text{TEV}}  \downarrow$} & \textbf{$\sigma_{\text{TEV}} \downarrow$}& \textbf{LAMBADA} $\downarrow$ & \textbf{WIKI2} $\downarrow$ & \textbf{WIKI103} $\downarrow$ &\textbf{PTB} $\downarrow$ & \textbf{1BW} $\downarrow$ \\
\toprule
GPT-2 &  &  0.0892 & 0.0125 & 79.60  & 44.74 & 54.53 & 53.05 & 59.28 \\
$\sigma$Reparam & 48 &  0.0879 & 0.0115  & 76.02 & 45.06 & 54.73 & 50.91 & 57.67 \\
MLRA &   &  \textbf{0.0875} & \textbf{0.0114} & \textbf{70.61} & \textbf{42.86} & \textbf{50.92} & \textbf{50.27} & \textbf{55.46} \\
\midrule
GPT-2 &  &  0.0872 &  0.0120 & 71.52 & 42.84 & 51.61 & 49.80 & 56.92 \\
$\sigma$Reparam & 96 &  0.0849 & 0.0113  & 70.39 & 42.34 & 50.23 & 49.53 & 55.72 \\
MLRA &  &  \textbf{0.0843} & \textbf{0.0110} & \textbf{62.31} & \textbf{39.44} & \textbf{46.22} & \textbf{44.17} & \textbf{51.56}  \\
\midrule
GPT-2 &  &  0.0875 & 0.0117 & 64.62 & 41.31 & 47.75 & 47.73 & 51.97 \\
$\sigma$Reparam & 192 & 0.0870 & 0.0112 & 59.86 & 39.06 & 44.13 & 43.79 & 48.51 \\
MLRA &  &  \textbf{0.0864} & \textbf{0.0104} & \textbf{53.69} & \textbf{35.39} & \textbf{44.17} & \textbf{41.14} & \textbf{45.03} \\
\bottomrule
\end{tabular}
\end{sc}
\end{small}
\end{center}
\caption{Zero-shot perplexity and token embedding variability (TEV) comparison between GPT-2, $\sigma$Reparam, and MLRA with varying number of layers. The bolded texts indicate the lowest $\mu_{\text{TEV}}$, $\sigma_{\text{TEV}}$ and perplexity across the model configurations with the same number of layers. 
The model dimension $d_{\text{model}}$ for GPT-2 and $\sigma$Reparam is set to 384, while the intermediate dimension $d_{\text{rank}}$ of MLRA is configured to 192. MLRA demonstrates both the lowest $\mu_{\text{TEV}}$ and $\sigma_{\text{TEV}}$ and perplexity, implying MLRA leads to the best pre-training stability and performance.}
\label{table:clm_merged}
\vskip -0.1in
\end{table*}

\section{Experiments}\label{sec:experiments}

We demonstrate the significance of TEV in Section \ref{subsec:design} and the effectiveness of MLRA on pre-training stability and performance in Section \ref{subsec:results}.

\subsection{Experimental Design}\label{subsec:design}

\paragraph{Baseline} We pre-train GPT-2 \citep{radford2019language} from scratch with three different methods: 1) conventional architecture (GPT-2), 2) $\sigma$Reparam \citep{zhai2023stabilizing}, and 3) MLRA. All the pre-training configurations, including learning rate and number of parameters, are fixed over methods. Further details can be found in Appendix \ref{apdx:experiments}.

\paragraph{Datasets} We pre-train each model using WebText \cite{radford2019language} and evaluate the downstream performances on Lambada \citep{paperno2016lambada}, Wikitext-2 \citep{merity2016pointer}, Wikitext-103 \cite{wikitext103}, Penn Tree Bank (PTB) \citep{10.5555/972470.972475}, and 1th  Billion Word Benchmark (1BW) \citep{chelba2014billion} datasets.

\subsection{Results}\label{subsec:results}
\paragraph{Pre-training stability} In Figure \ref{fig:grad_var}, MLRA has the lowest gradient variance in all configurations when pre-trained on the first one billion tokens. As models deepen, the gradient variance gap between baselines and MLRA is increasingly pronounced. A significant spike in gradient variance around 600M tokens across all configurations suggests high optimization difficulty \cite{faghri2020study}. We excluded the result of GPT-2 with 192 layers as it failed five times during pre-training, showing the pre-training instability of GPT-2 as it gets deeper Deeper model pre-training is unstable \cite{wang2022deepnetscalingtransformers1000, wang2022foundation} due to shattered gradients resembling white noise \cite{balduzzi2018shatteredgradientsproblemresnets}.

\begin{figure}
    \centering
    \includegraphics[width=\columnwidth]{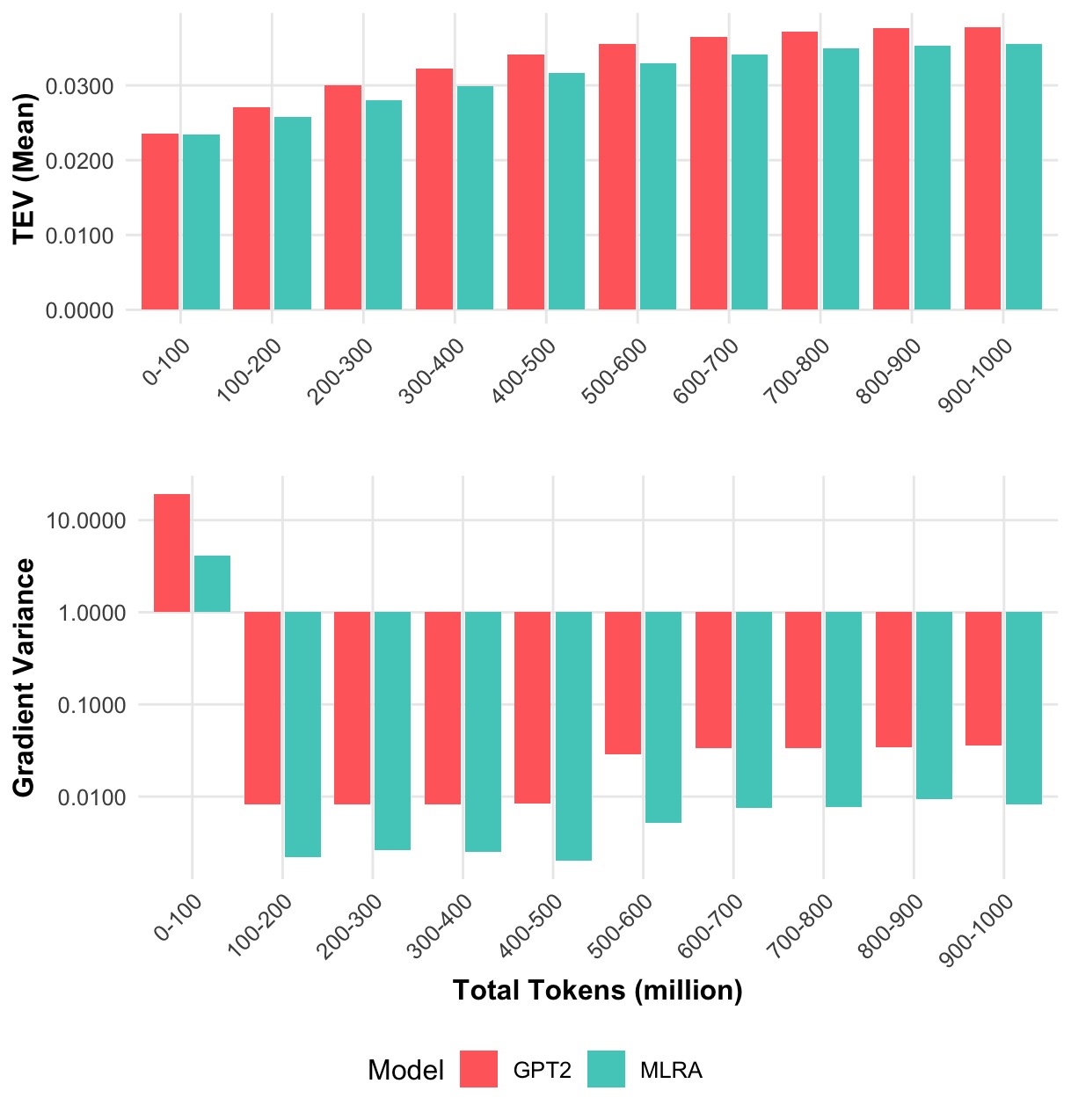}
    \caption{$\mu_{\text{TEV}}$ (top) and gradient variance (bottom) during the pre-training of both GPT-2 and MLRA, each with 48 layers, over the course of 1 billion tokens. For both settings, $\mu_{\text{TEV}}$ and gradient variance imply identical trends over the pre-training procedure.}
    \label{fig:enter-label}
\end{figure}
\paragraph{Token Embedding Variability} Aligned with the results in Figure \ref{fig:grad_var}, MLRA \emph{consistently} exhibits lower $\mu_{\text{TEV}}$ and $\sigma_{\text{TEV}}$ compared to GPT-2 and $\sigma$Reparam in Table \ref{table:clm_merged}. Moreover, $\mu_{\text{TEV}}$ and $\sigma_{\text{TEV}}$ for 192 layers are higher than those for 96 layers, indicating increased gradient variance at this deeper layer, as shown in Figure \ref{fig:grad_var}.
We further study the correlation between gradient variance and $\mu_{\text{TEV}}$ over 1 billion tokens. Figure \ref{fig:enter-label} shows that higher gradient variance corresponds to a higher $\mu_{\text{TEV}}$, with the rate of increase in $\mu_{\text{TEV}}$ depending on the magnitude of gradient variance.

\paragraph{Perplexity performance}
As can be observed in Table~\ref{table:clm_merged}, the zero-shot performances of MLRA are significantly improved compared to baselines across different numbers of layers and datasets on which these models are not fine-tuned. We also achieve better zero-shot perplexity results than $\sigma$Reparam \cite{zhai2023stabilizing}, the current state-of-the-art model that alleviates the attention entropy collapse problem. Furthermore, the perplexity gap of MLRA becomes much larger than the vanilla counterparts as the number of layers increases. These findings empirically prove the efficacy and depth-scalability of the proposed method.

\section{Conclusion}

This paper shows that Token Embedding Variability (TEV) can be used as a simple and efficient proxy for pre-training stability, avoiding the high cost of monitoring gradient variance. Theoretical analysis reveals that factorized multi-head attention projection matrices (\textit{i.e.}, MLRA) reduce gradient explosion. Empirically, MLRA lowers TEV mean and variance, improves stability, and outperforms GPT-2 and $\sigma$Reparam in reducing zero-shot perplexity, particularly in deeper models.

\section*{Limitations}

While we conducted a controlled study of the pre-training stability and token embedding variability (TEV) as a proxy by pre-training GPT-2 from scratch, the scale of the base model was limited to a maximum of 1.5B parameters. We also compare the performance and stability with a single pre-training corpus, the WebText. Therefore, the scalability of MLRA and TEV as a pre-training stability proxy will be further studied across a larger range of scales, 7B, for instance.

\bibliography{custom}

\appendix
\appendix
\onecolumn

\section{Related Works}
\label{apdx:related}

\paragraph{Training instability in LLMs} Modern LLMs, such as the GPT series \cite{radford2018improving, radford2019language, brown2020language} and llama series \cite{touvron2023llama, touvron2023llama2} frequently use pre-layer normalization (pre-LN), which normalizes inputs instead of outputs \cite{zhai2023stabilizing}. Pre-LN increases the standard deviation of hidden representation in upper layers, preserving unique data features and preventing token embeddings from becoming too similar \cite{brunner2019identifiability}. However, it can cause gradient explosion in shallow layers, where gradients from shallower layers grow disproportionately larger than those from deeper layers, affecting training stability \cite{shleifer2021normformer, takase2024spike}. \citealp{takase2024spike} shows that in pre-LN settings, sub-component norms grow exponentially when standard deviations exceed 1, which is a common issue with typical initialization. To address this, methods like sub-LayerNorm \citep{shleifer2021normformer,wang2022foundation} and sigma reparameterization ($\sigma$Reparam) \cite{zhai2023stabilizing}, which scales weights by their spectral norms, have been developed to enhance stability. Scaled initialization which scales down the initial weight values \cite{shoeybi2020megatronlm, scao2022language} also helps mitigate gradient spikes during pre-training.

\paragraph{Low-rank pre-training} A plethora of literature regarding low-rank training has been conducted in the domains of convolution neural network (CNN) compression, regularization, and the pursuit of efficient training and inference \cite{low-rank_NN, jaderberg2014speeding, sui2022elrt, low-rank_lottery_ticket, winata2020lightweight}. Nevertheless, most of these methods are tailored exclusively for CNNs and have yet to undergo assessment on large-scale Transformers \cite{vaswani2023attention}, which could significantly benefit from efficient training due to the large scale of language models.

Recently, methods like ReLoRA \cite{lialin2023stack} and InRank \cite{zhao2023inrank} have adopted an approach that starts training with full-rank matrices and then transitions to low-rank training. These studies suggest that the intrinsic rank of Transformers decreases as training progresses \cite{aghajanyan2020intrinsic, hu2021lora}. In earlier phases, full-rank matrices are used to stabilize training before switching to low-rank matrices after a few initial steps.

\section{Mean of Token Embedding in pre-trained LLM}
\label{apdx:mean}

\begin{figure*}[t!]
    \centering
    \includegraphics[width=\textwidth]{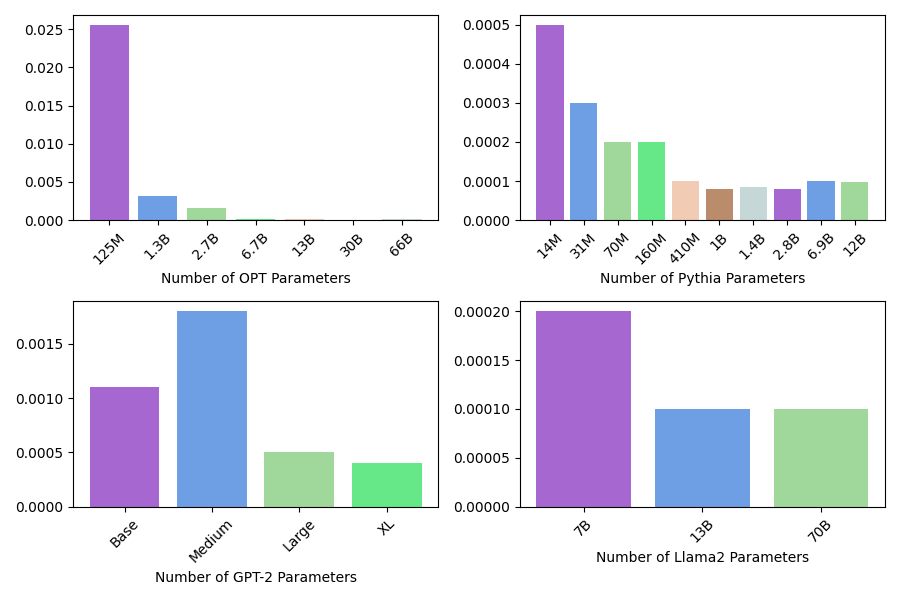}
    \caption{Row-wise average of the absolute mean value of $|V|$ token embeddings in the token embedding layer $\mathbf{E} \in \mathbb{R}^{|V| \times d_{\text{model}}}$ across OPT \cite{zhang2022opt}, Pythia \cite{biderman2023pythia}, Llama-2 \cite{touvron2023llama2} and GPT-2 \cite{radford2019language}. $\mathbf{E}$ in pre-trained checkpoint remains centered around zero.}
    \label{fig:embed_mean}
\end{figure*}

Figure \ref{fig:embed_mean} illustrates that the row-wise average of the absolute mean value of $|V|$ token embeddings in the token embedding layer $\mathbf{E} \in \mathbb{R}^{|V| \times d_{\text{model}}}$ across OPT \cite{zhang2022opt}, Pythia \cite{biderman2023pythia}, Llama-2 \cite{touvron2023llama2} and GPT-2 \cite{radford2019language} remains centered around zero after pre-training. One possible conjecture on this phenomenon is that pre-LN \cite{prelayernorm} introduces layer normalization before the logits, resulting in a similar effect as logit normalization \cite{wei2022mitigatingneuralnetworkoverconfidence}. This process enforces a constant vector norm on the logits during training, helping to alleviate the issue of overconfidence (\emph{i.e.} unusually high softmax confidences, even when the inputs are significantly different from the training data). Additionally, a slight negative correlation between model size and the embedding mean is observed, warranting further investigation.

\section{Variance of Kaiming Uniform Initialization}
\label{apdx:kaiming}

The Kaiming uniform initialization is defined by the following distribution:

\[
w \sim \mathcal{U}\left(-\sqrt{\frac{6}{n \cdot (1 + a^2)}}, \sqrt{\frac{6}{n \cdot (1 + a^2)}}\right)
\]

where \( \mathcal{U}(a, b) \) denotes the uniform distribution between \(a\) and \(b\). \( n \) is the number of input units in the weight tensor. \( a \) is a scaling parameter, given as \( \sqrt{5} \) in this case.

Given \( a = \sqrt{5} \), we have \( a^2 = 5 \). Therefore, the range of the uniform distribution becomes:

\[
w \sim \mathcal{U}\left(-\sqrt{\frac{6}{n \cdot 6}}, \sqrt{\frac{6}{n \cdot 6}}\right) = \mathcal{U}\left(-\sqrt{\frac{1}{n}}, \sqrt{\frac{1}{n}}\right)
\]

The variance of a uniform distribution \(\mathcal{U}(a, b)\) is given by:

\[
\sigma^2(\mathcal{U}(a, b)) = \frac{(b - a)^2}{12}
\]

For our distribution:

\[
a = -\sqrt{\frac{1}{n}}, \quad b = \sqrt{\frac{1}{n}}
\]

The range width \( b - a \) is:

\[
b - a = \sqrt{\frac{1}{n}} - \left(-\sqrt{\frac{1}{n}}\right) = 2\sqrt{\frac{1}{n}}
\]

Thus, the variance is:

\[
\sigma^2(w) = \frac{\left(2\sqrt{\frac{1}{n}}\right)^2}{12} = \frac{4 \cdot \frac{1}{n}}{12} = \frac{1}{3n}
\]

\section{Further Extension of \ref{sec:theory}}
\label{apdx:extension}

In this section, we calculate output representation variance after a single attention head and self-attention. To simplify the equation, let $\text{softmax}\left(\frac{X_t W_{Q_i} (X_t W_{K_i})^T}{\sqrt{d_{\text{head}}}}\right)$ be \( A \). For the simplicity of calculation, we assume $d_{\text{model}}$ = $d_{\text{head}}$, which is a single-head attention.
Because \(X\) is layer-normalized input, $\sigma^2(\text A X_t)$ reaches the maximum value of 1 when the result of \(\text A\) is a one hot vector. Thus, the upper-bound variance of each head $i$ and attention in the initialization stage of MLRA are as follows: 
\begin{equation}
\begin{split}
    \sigma^2(\text{head}(X_t)) &= \sigma^2(\text A X_t) \cdot d_{\text{r}} \cdot d_{\text{model}} \cdot \sigma^2(W^{U}) \cdot \sigma^2(W^{D}) \\ %
    &= \sigma^2(\text A X_t) \cdot d_{\text{r}} \cdot d_{\text{model}} \cdot \frac{1}{3d_{\text{model}}} \cdot \frac{1}{3d_{\text{r}}} \\
    &< \frac{1}{9}\\
\end{split}
\label{eq:head_var}
\end{equation}
\begin{equation}
\begin{split}
    \sigma^2(\text{Attention}(X_t)) &= \sigma^2(\text{head}_i(X_t)) \cdot d_{\text{model}} \cdot \sigma^2(W_O) \\
    &= \sigma^2(\text{head}_{i}(X_t)) \cdot d_{\text{model}} \cdot \frac{1}{3d_{\text{model}}}  \\
    & < \frac{1}{27}
\end{split}
\label{eq:attention_var}
\end{equation}

where $\sigma^2(W_O) \in \mathbb{R}^{d_{\text{model}} \times d_{\text{model}}}$. The calculation shows that attention weights $W \in \mathbb{R}^{d_{\text{model}} \times d_{\text{model}}}$ have a variance upper bound of $\frac{1}{9}$ per head and $\frac{1}{27}$ for the entire module. In contrast, MLRA's variance upper bound is one-third lower under Kaiming uniform initialization.

\section{Implementation Details}
\label{apdx:experiments}

\paragraph{Configuration} We measure TEV and apply MLRA to the widely adopted GPT-2 language model configuration \cite{radford2019language} Specifically, we pre-train GPT-2 \cite{radford2019language}, $\sigma$Reparam \cite{zhai2023stabilizing}, and MLRA with hidden dimensions 384 and depth layers of 48, 96, 192 using the WebText dataset \cite{radford2019language}, where total number of token is $5.5$B. Each model is trained with 4 epochs with the casual language modeling objective, as a recent study experimentally shows that repeating data more than 4 times in a decoder-only model with a data-constrained regime is computationally inefficient \cite{muennighoff2023scaling}. We set a batch size of 512 and a learning rate of 1e-3 for the base model. All model types in this paper follow the same training configuration for consistency.

\paragraph{Model assessment} For evaluation of GPT-2 models on the upstream language modeling tasks, we follow conventions in language modeling and report the perplexity, which measures average log probabilities of each sentence token predictions in an autoregressive way \cite{radford2019language}:
\begin{equation}
    \text{PPL}(W) = \exp \left(- \frac{1}{N} \sum^N_{t=1} \log P(x_t|x_{<t};\Theta)\right)\label{eq:ppl}
\end{equation}
where $\mathbf{x}=\{x_1, x_2. ..., x_N\}$ are the set of $N$ tokens.

\end{document}